\documentclass[10pt,twocolumn,letterpaper]{article}

\usepackage{cvpr}
\usepackage{times}
\usepackage{epsfig}
\usepackage{graphicx}
\usepackage{amsmath}
\usepackage{amssymb}

\usepackage{multirow}
\usepackage{subfigure}
\usepackage{graphicx}
\usepackage{capt-of}

\usepackage[pagebackref=true,breaklinks=true,letterpaper=true,colorlinks,bookmarks=false]{hyperref}

\cvprfinalcopy 

\def\cvprPaperID{1793} 

\ifcvprfinal\fi
\begin{document}

\title{Learning Disentangling and Fusing Networks for Face Completion Under Structured Occlusions}

\author{
Zhihang Li \quad Yibo Hu \quad Ran He \\
National Laboratory of Pattern Recognition, CASIA\\
Center for Research on Intelligent Perception and Computing, CASIA\\
University of Chinese Academy of Sciences, Beijing, 100049, China\\
{\tt\small \{zhihang.li, rhe\}@nlpr.ia.ac.cn}, 
{\tt\small yibo.hu@cripac.ia.ac.cn}
}

\maketitle

\begin{abstract}
   Face completion aims to generate semantically new pixels for missing facial components. It is a challenging generative task due to large variations of face appearance. This paper studies generative face completion under structured occlusions. We treat the face completion and corruption as disentangling and fusing processes of clean faces and occlusions, and propose a jointly disentangling and fusing Generative Adversarial Network (DF-GAN). First, three domains are constructed, corresponding to the distributions of occluded faces, clean faces and structured occlusions. The disentangling and fusing processes are formulated as the transformations between the three domains. Then the disentangling and fusing networks are built to learn the transformations from unpaired data, where the encoder-decoder structure is adopted and allows DF-GAN to simulate structure occlusions by modifying the latent representations. Finally, the disentangling and fusing processes are unified into a dual learning framework along with an adversarial strategy. The proposed method is evaluated on Meshface verification problem. Experimental results on four Meshface databases demonstrate the effectiveness of our proposed method for the face completion under structured occlusions.
\end{abstract}

\section{Introduction}

Face completion \cite{GFC-CVPR-2017} refers to the task of filling the missing or occluded regions with semantically consistent contents in face images. Since face images have large appearance variations and possess high-level identity information, this task is more difficult than traditional image completion \cite{afonso2011augmented,hu2013fast,barnes2009patchmatch,efros1999texture,hays2007scene} which pays more attention to visually effects. Face completion is quite helpful for a wide spectrum of downstream applications, such as face alignment \cite{zhang2016demeshnet} and face verification \cite{zhang2016multi}. However, recovering a clean face from an occluded one is an ill-posed or under-defined problem where many clean images may correspond to one occluded input. Therefore, face completion remains a challenge in computer vision.

There have been several attempts for face completion, however, they usually focus on addressing general image completion. Existing image completion methods can be broadly categorized into three groups. The first one is based on prior knowledge of images such as smoothness \cite{afonso2011augmented} or low rank \cite{hu2013fast,yang2017nuclear}, which is useful to fill small missing or some less-textured regions. However, these prior assumptions are hard to satisfy if the missing regions contain complicated textures. The second group adopts a copy-and-paste strategy \cite{barnes2009patchmatch,efros1999texture,hays2007scene} that searches the most similar patch in the known regions of the image \cite{barnes2009patchmatch} or an extra database \cite{hays2007scene}, and then pastes it into the missing areas. Although the copy-and-paste based methods are simple and efficient, it cannot work well when the missing areas are not well represented by any images in the external database. Recently, learning based methods have achieved promising results \cite{liu2016learning,pathak2016context,yeh2017semantic,yang2016high,GFC-CVPR-2017,huang2017wavelet}, where a learnable model such as a neural network is trained on a training set. Inference is conducted to fill the missing regions after the model is well trained. Although some learning based methods have been proposed to address different types of missing regions, most of them utilize the known parts of the occluded images but rarely consider the structure of the occlusions. 
\begin{figure*}[htb]
\vspace{-3pt}
	\center
    \includegraphics[width=0.8\textwidth]{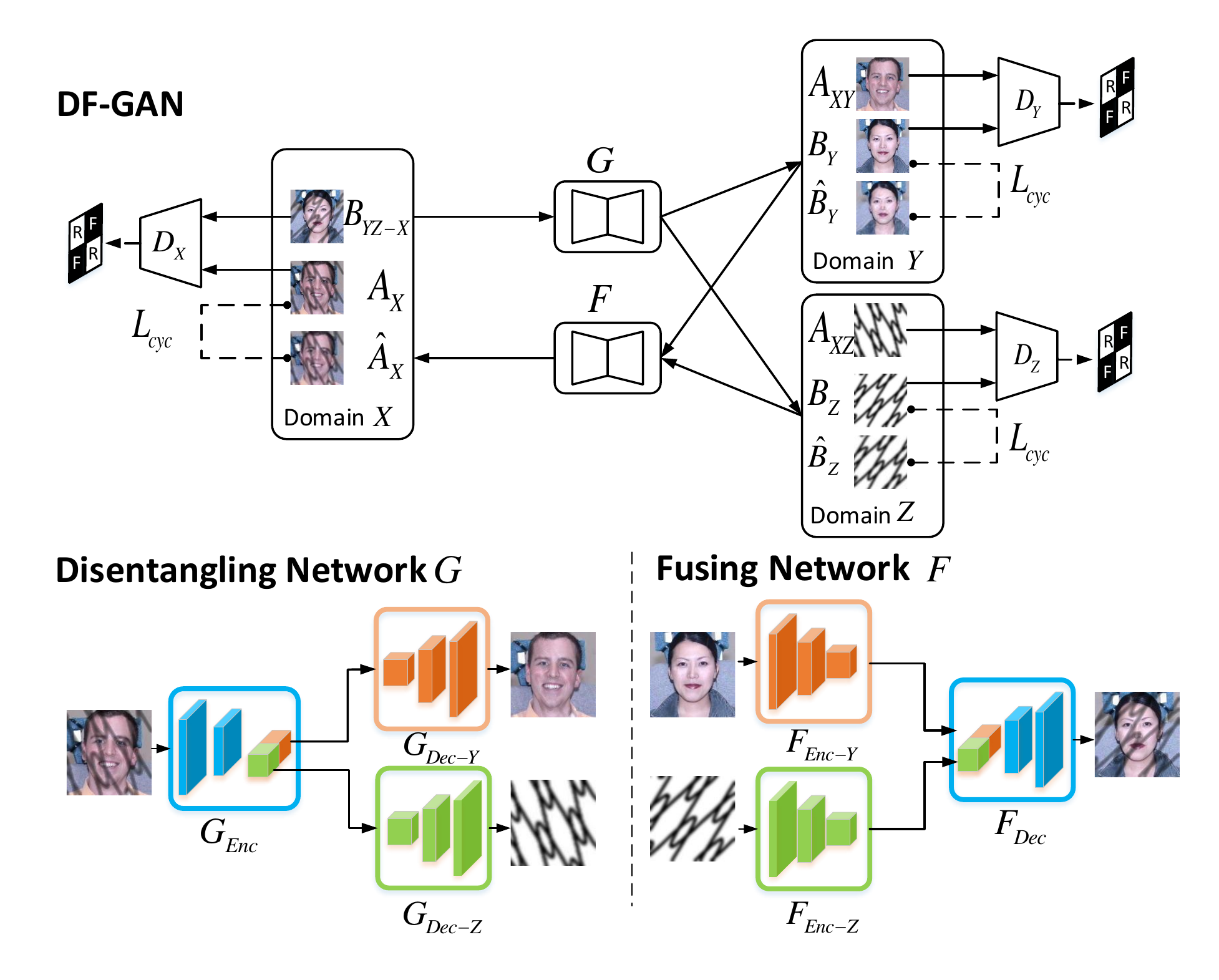}\captionof{figure}{The framework of our proposed DF-GAN. It consists of two generators: $G$ and $F$, and three domain-specific discriminators ${D_X},{D_Y},{D_Z}$. For a real MeshFace ${A_X}$ in domain $X$, $G$ transforms ${A_X}$ into a fake clean face ${A_{XY}}$ in domain $Y$ and a fake mesh ${A_{XZ}}$ in domain $Z$. Then ${A_{XY}}$ and ${A_{XZ}}$ are transformed back to domain $X$ and reconstruct ${\hat A_X}$ by $F$. Similarly, a real clean face ${B_Y}$ and mesh ${B_Z}$ are fed into $F$ to synthesize a fake MeshFace ${B_{YZ - X}}$ in domain $X$ that is employed to reconstruct ${\hat B_Y}$ and ${\hat B_Z}$ by $G$.}
    \label{fig::framework}
   \vspace{-3pt}
\end{figure*}
Besides, all these works require the availability of paired samples. However, it is difficult to simultaneously collect the occluded images and their corresponding clean images in some real application scenarios.

This paper focuses on the face completion problem with structured occlusions and explicitly models the structured occlusions to boost performance. We propose a simultaneously disentangling and fusing Generative Adversarial Network (DF-GAN) for jointly face completion and occlusions modeling without requiring the paired face images for training (as shown in Figure \ref{fig::framework}). Different from previous methods that view image completion as a missing region filling problem, the face completion and corruption are treated as disentangling and fusing processes of clean faces and occlusions in our model. Specifically, three domains are firstly constructed, corresponding to the distributions of occluded faces, clean faces and structured occlusions. Then, the disentangling and fusing networks are created to learn the transformations between the three domains. In disentangling network, an occluded face is encoded to the disentangled representations by an encoder, and two decoders then generate the corresponding clean face and occlusion respectively. For any samples in the domains of clean faces and occlusions, the fusing network simply concatenates their latent representations and then synthesizes the corresponding occluded face. The disentangling and fusing processes can be unified into a dual learning framework, which allows images to be transferred from one domain to another. Finally, three domain-specific discriminators are imposed to compete with generators, which are trained by an adversarial strategy. Our DF-GAN is evaluated on Meshface verification problem \cite{zhang2016multi,zhang2016demeshnet}, where MeshFace refers to a type of face images occluded by random mesh-like occlusions that have been widely used to protect private information from abusing \cite{zhang2016multi,zhang2016demeshnet}. Experimental results on four Meshface databases demonstrate the effectiveness of our proposed method for the face completion under structured occlusions.

The main contributions of this paper are:

1) We propose a novel view of face completion where completion and corruption are treated as disentangling and fusing processes of clean faces and occlusions. These processes are formulated as the transformations between three domains.

2) In contrast to existing face completion methods that rarely consider the structure of occlusions, our method explicitly models the structured occlusions. We create the disentangling network and fusing network that adopt an encoder-decoder structure to learn representations of the three domains, which allows simulating structure occlusions by modifying the latent representations.

3) Without requiring the paired face images for training, we unify the face completion and corruption into a dual learning framework, which is effective to deal with unpaired data. Quantitative and visual evaluations on both controlled and in-the-wild databases demonstrate the effectiveness of DF-GAN for the face completion under structured occlusions.

\section{Related Work}
Since our method is related to GAN and image completion, the relevant works are briefly reviewed.

\noindent \textbf{Generative Adversarial Network.} Goodfellow et al. \cite{goodfellow2014generative} proposed Generative Adversarial Network (GAN) to learn a generative model by an adversarial process. It consists of a generator and a discriminator, which are alternatively trained in a minimax two-player game. GAN has been improved with a deeper convolutional architecture (DCGAN) by Radford et al. \cite{radford2015unsupervised}. Additionally, the conditional version of GAN (cGAN) \cite{Mirza2014Conditional} was introduced, which is able to generate images under an image, a class label or landmarks. To deal with unpair data, CycleGAN \cite{CycleGAN2017}, DualGAN \cite{Yi_2017_ICCV} and DiscoGAN \cite{pmlr-v70-kim17a} were proposed to learn image-to-image translation from a source domain to a target domain. However, without considering the structure of faces and occlusions, they can not be applied well to face completion. 

\noindent \textbf{Image completion.} Massive algorithms have been proposed to deal with image completion, including prior knowledge based methods \cite{afonso2011augmented,hu2013fast,he2014half}, copy-and-paste based methods \cite{barnes2009patchmatch,efros1999texture,hays2007scene} and learning based methods \cite{xie2012image,ren2015sh,liu2016learning,pathak2016context,yeh2017semantic,yang2016high,GFC-CVPR-2017,huang2017wavelet}. An early completion method \cite{hu2013fast} exploits the low-rank structure in natural images and translates the image completion problem as an optimization of matrix completion, which is effective to small and homogeneous regions. The copy-and-paste based method is computationally efficient, which searches the most similar patch in a single image \cite{barnes2009patchmatch} or an external database \cite{hays2007scene}. However, it is more likely to fail when the similar patch cannot be found in the extra database. The learning based methods, especially for deep learning, spring up and have produced compelling visual results. Phatak et.al \cite{pathak2016context} proposed the Context Encoders, which is an encoder-decoder structure and trained with $L2$ and adversarial losses between the predicted and real missing regions. A high-resolution image inpainting method was proposed by Yang et al. \cite{yang2016high}, where they design a content network and a textural network to learn global structure and local texture simultaneously. Two most relevant works to face completion are \cite{GFC-CVPR-2017, yeh2017semantic}. \cite{GFC-CVPR-2017} utilizes two adversarial losses to ensure the local-global contents consistency and introduces a semantic parsing network to preserve the semantic information. \cite{ yeh2017semantic} proposed a manifold search method, where the closest mapping on the latent manifold is found and the missing content is reconstructed by a generative model. However, most of the existing methods mentioned above rarely consider the structure of occlusions. Furthermore, they almost resort to paired face images for training.

\section{The Proposed Method}

In this section, we aim to address the issue of face completion under structured occlusions. As shown in Figure \ref{fig::framework}, our method contains three parts: 1) the disentangling network $G$; 2) the fusing network $F$; 3) three domain-specific discriminative networks ${D_X},{D_Y}$ and ${D_Z}$. 

The disentangling network $G$ and the fusing network $F$ can be seen as two generators to accomplish a dual task. To separate clean faces from occlusions, the generator $G$ achieves a transformation from domain $X$ to domain $Y$ and domain $Z$, i.e., $X \to Y,Z$. The generator $F$ performs a reverse task, which fuses samples in domain $Y$ and domain $Z$ to synthesize occluded faces, i.e., $Y,Z \to X$. $G$ and $F$ compose a closed loop, which is addressed by dual learning \cite{he2016dual}. DF-GAN incorporates three domain-specific discriminative networks ${D_X},{D_Y}$ and ${D_Z}$, which are the key to capturing the distributions of real data in three domains.

In the following sub-sections, the three components are described in detail. Then we derive the adversarial framework of DF-GAN.
\subsection{The Disentangling Network}

The disentangling network $G$ imposes an exclusiveness constraint on latent space to separate clean faces from occlusions in MeshFaces. The detailed structure of $G$ is shown in Figure \ref{fig::framework}, which consists of one encoder and two decoders. The encoder ${G_{Enc}}$ maps the input occluded face $x$ to a latent representation, i.e., ${G_{Enc}}(x)$. According to the exclusiveness constraint, the latent space is factorized into two parts ${G_{Enc{\rm{|}}Y}}(x)$, ${G_{Enc{\rm{|}}Z}}(x)$ for the clean face and the occlusion respectively. It requires that the representations of ${G_{Enc{\rm{|}}Y}}(x)$ and ${G_{Enc{\rm{|}}Z}}(x)$ shall be exclusively captured and not interweave with each other. To complete the transformation to domain $Y$ and domain $Z$, two decoders ${G_{Dec - Y}}$, ${G_{Dec - Z}}$ generate the corresponding clean face $\hat y$ and occlusion $\hat z$. The separated clean face from occlusion by $G = \{ {G_{Enc}},{G_{Dec - Y}},{G_{Dec - Z}}\} $ are formulated as:
\begin{equation}
\begin{array}{l}
\hat y = {G_{Dec - Y}}({G_{Enc|Y}}(x))\\
\hat z = {G_{Dec - Z}}({G_{Enc|Z}}(x))
\end{array}
\label{eq:1}
\end{equation}
where ${G_{Enc}}(x) = [{G_{Enc|Y}}(x),{G_{Enc|Z}}(x)]$ is the disentangled representation of $x$.

\subsection{The Fusing Network}

The objectives of the fusing network $F$ are twofold: 1) learn domain-specific representations for domain $Y$ and domain $Z$ respectively, and 2) fuse clean faces and occlusions to synthesize corresponding occluded face in domain $X$. As opposed to $G$, the generator $F$ contains two encoders ${F_{Enc - Y}},{F_{Enc - Z}}$ and one decoder ${F_{Dec}}$. Since our method fuses clean faces and occlusions in latent space instead of high-dimension pixel space, clean face $y$ and occlusion $z$ are firstly mapped to their latent representations ${F_{Enc - Y}}(y),{F_{Enc - Z}}(z)$ by the two encoders respectively. To transfer back to the domain $X$, the decoder fuses their representations and generates the corresponding occluded face $\hat x$. The encoder-decoder structure provides a convenient way to control the synthetic occluded faces. We can directly perform simple arithmetic operations in latent space to generate different occlusions. The synthetic image by the fusing generator $F = \{ {F_{Enc - Y}},{F_{Enc - Z}},{F_{Dec}}\} $ is defined as following:
\begin{equation}
\hat x = {F_{Dec}}({F_{Enc - Y}}(y),{F_{Enc - Z}}(z))
\end{equation}
Where $\hat x$ is the occluded image generated by $F$ in domain $X$. We simply concatenate the latent representations of clean face and occlusion in our implementation.

\subsection{The Domain-specific Discriminative Networks }

Three domain-specific discriminators ${\rm{\{ }}{D_X},{D_Y},{D_Z}{\rm{\} }}$ aim to incorporate priors about the structure of clean faces and occlusions. Different from DR-GAN \cite{tran2017disentangled} that resorts to the paired one-to-one label (identity and pose are required for each image) to disentangle attribute (pose) in feature space, our model has no such paired data to decouple features. Thus, the naive combination of disentangling network and fusing network by dual learning is insufficient. In contrast, it is easy for person to separate clean faces from occlusions without the pairwise correspondence by exploiting some priors. To incorporate such priors into our model, we design three domain-specific discriminators to judge whether the generators have learned the priors. If the discriminators fail to distinguish between the real sample and the generated ones, the generators have successfully learned such priors of faces and occlusions. Therefore, three discriminators implicitly facilitate the disentangled representations of clean faces and occlusions in latent space.

\subsection{Adversarial Training}
We construct an adversarial game between the generators and the discriminators. Specifically, the generators attempt to generate samples as real as possible to fool the discriminators, while the discriminators are trained to distinguish between the real samples $x,y,z$ and the generated ones $\hat x, \hat y, \hat z$. Formally, three domains $X$, $Y$ and $Z$ are constructed corresponding to the occluded faces, clean faces and occlusions, where the discriminators ${\rm{\{ }}{D_X},{D_Y},{D_Z}{\rm{\} }}$ for each domain are imposed to learn the priors about the structure of faces and occlusions. To learn the disentangled features and the transformation from domain $X$ to domain $Y$ and domain $Z$, the disentangling network $G$ is created. In addition, the fusing network $F$, as a dual task, achieves the fusion of domain $Y$ and domain $Z$, and synthesizes the occluded faces. Three discriminators learn to differentiate real images from the synthesized ones on each domain, while two generators $G,F$ try to generate realistic visual images by minimizing the discriminators' chance of correctly telling apart the sample source. Following the original GAN \cite{goodfellow2014generative} that optimizes over binary probability distance, the objective of the adversarial framework is thus written as:
\begin{equation}
\setlength{\jot}{5pt}
\begin{aligned}
&\mathop {\min }\limits_{G,F} \mathop {\max }\limits_{{D_X},{D_Y},{D_Z}} {L_{GAN}}(G,F,{D_X},{D_Y},{D_Z}) \\
&= {E_{x\sim{P_{data}}(x)}}[\log (1 - {D_Y}(\hat y)) + \log (1 - {D_Z}(\hat z))]\\
&+ {E_{y\sim{P_{data}}(y),z\sim{P_{data}}(z)}}[\log (1 - {D_X}(F(y,z)))] \\
&+ {E_{x\sim{P_{data}}(x)}}[\log {D_X}(x)] + {E_{y\sim{P_{data}}(y)}}[\log {D_Y}(y)] \\
&+ {E_{z\sim{P_{data}}(z)}}[\log {D_Z}(z)] 
\end{aligned}
\end{equation}
Where ${P_{data}}(x),{P_{data}}(y),{P_{data}}(z)$ denote the true data distributions of the three domains, respectively. $\hat y,\hat z$ are the synthesized clean face and occlusion by $G$ in domain $Y$ and domain $Z$ as in Eq.(\ref{eq:1})

In our model, the disentangling generator and the fusing generator compose a closed loop. Inspired by dual learning \cite{he2016dual}, we extend the cycle loss on two domains \cite{CycleGAN2017,Yi_2017_ICCV,pmlr-v70-kim17a} to three domains, which is effective to deal with unpaired data. Mathematically, the cycle consistency loss on three domains is written as:
\begin{equation}
\begin{split}
{L_{cyc}}(G,F) = {E_{x\sim{P_{data}}(x)}}[&||F(G(x)) - x||] \\
+ {E_{y\sim{P_{data}}(y),z\sim{P_{data}}(z)}}[&||G(F(y,z)) - y|| \\
+ &||G(F(y,z)) - z||]
\end{split}
\end{equation}
Where ${L_{cyc}}(G,F)$ ensures the consistency of domain transformation. 

To sum up, the goal of our approach is to optimize the following loss function:
\begin{equation}
\mathop {\min }\limits_{G,F} \mathop {\max }\limits_{{D_X},{D_Y},{D_Z}} {L_{GAN}}(G,F,{D_X},{D_Y},{D_Z}) + \lambda {L_{cyc}}(G,F)
\end{equation}
Where $\lambda $ is the trade-off parameter. The optimization of the generators and the discriminators is conducted in an alternatively two-player min-max game manner.

\section{Experiments}
\begin{table*}[htbp]
  \centering
  \scalebox{0.95}{
    \begin{tabular}{c|c|c|c|c||c|c|c|c||c|c|c|c}
    \hline
    \multirow{2}[0]{*}{Method} & \multicolumn{4}{c||}{TPR@FPR=1\%} & \multicolumn{4}{c||}{TPR@FPR=0.1\%} & \multicolumn{4}{c}{TPR@FPR=0.01\%}  \\ \cline{2-13}
      & AR & MultiPIE & FERET & LFW & AR & MultiPIE & FERET & LFW & AR & MultiPIE & FERET & LFW \\
    \hline \hline
    Corrupted  & 84.19 & 51.86 & 94.82 & 75.13   & 67.47 & 25.54 & 86.04 & 52.90   & 52.33 & 10.63 & 70.04 & 45.87 \\ \hline
    Clean      & \textbf{\underline{97.44}} & \textbf{\underline{87.27}} & \textbf{\underline{100.00}} & \textbf{\underline{93.43}}           & \textbf{\underline{91.04}} & \textbf{\underline{74.41}} & \textbf{\underline{100.00}} & \textbf{\underline{90.67}}           & \textbf{\underline{85.68}} & \textbf{\underline{66.67}} & \textbf{\underline{99.57}} & \textbf{\underline{87.52}}\\ \hline \hline
    CycleGAN   & 87.11 & 76.19 & 97.03 & 83.53   & 73.91 & 55.17 & 95.57 & 72.13   & 51.41 & 34.65 & 87.72 & 62.23 \\ \hline
    DF-GAN     & \textbf{93.04} & \textbf{81.84} & \textbf{99.99} & \textbf{89.67}     & \textbf{84.81} & \textbf{65.80} & \textbf{99.81} & \textbf{84.33}     & \textbf{71.46} & \textbf{48.03} & \textbf{98.77} & \textbf{74.13}  \\ \hline
    \end{tabular}%
    }
    \caption{Verification performance on AR, MultiPIE, FERET and LFW datasets}
  \label{tab::TPR_FPR}%
\end{table*}%

\begin{table*}[htbp]
  \centering
    \begin{tabular}{c|c|c|c|c||c|c|c|c}
    \hline
    \multirow{2}[0]{*}{Method} & \multicolumn{4}{c||}{PSNR} & \multicolumn{4}{c}{SSIM}  \\ \cline{2-9}
      & AR & MultiPIE & FERET & LFW & AR & MultiPIE & FERET & LFW  \\
    \hline \hline
    Corrupted  & 17.79 & 16.85 & 19.99 & 18.23   & 0.8424 & 0.7408 & 0.8425 & 0.8445   \\ \hline \hline
    CycleGAN   & 19.07 & 25.61 & 24.78 & 20.08   & 0.8128 & 0.8977 & 0.8854 & 0.8562   \\ \hline
    DF-GAN     & \textbf{23.85} & \textbf{28.21} & \textbf{28.15} & \textbf{23.18}     & \textbf{0.9168} & \textbf{0.9176} & \textbf{0.9310} & \textbf{0.8690}    \\ \hline
    \end{tabular}%
    \caption{Completion results on AR, MultiPIE, FERET and LFW datasets}
  \label{tab::PSNR_SSIM}%
\end{table*}%

In this section, the proposed DF-GAN is evaluated on four MeshFace datasets, including three datasets under controlled conditions and one dataset in the wild. MeshFace is often corrupted by mesh-like occlusions that have random position, width and transparency \cite{zhang2016multi,zhang2016demeshnet}. Particularly, MeshFace completion requires both generating clean faces and improving verification performance. Hence, MeshFace potentially provides a good platform to evaluate different face completion methods for structured occlusions. We first introduce the used datasets and testing protocols. The baseline methods and implementation details are then specified. At last, a comprehensive experimental analysis is conducted on synthesis visual results and quantitative face verification results.

\subsection{Datasets and Protocols}
\textbf{The AR face database} \cite{martinez1998ar}. It contains over 4,000 color images of 126 people. Each people has several frontal view faces with different facial expressions, lighting conditions and occlusions. In addition, 130 landmarks for each people are provided. In our experiments, the frontal faces without occlusions but with landmarks are selected. Finally, we obtain 112 people with 895 face images including 56 people for training and 56 people for testing.

\textbf{The CMU MultiPIE face database} \cite{gross2010multi}. This dataset is the largest database for evaluating face recognition under different poses, illumination conditions and expressions. Moreover, MultiPIE provides 68 landmarks for each image. We select images with frontal view and balanced illumination in our experiment, resulting in 337 subjects with 2,403 images. The first 100 subjects are for testing and the rest 237 for training. 

\textbf{The Color FERET database} \cite{phillips2000feret}. Images on the Color FERET database are taken under controlled condition. The database contains images of 1,199 different individuals with different poses. People with `fa' and `fb' frontal images are chosen for our experiment, which results in 966 subjects. We split the database into a training set of 666 subjects and a testing set of 300 subjects. 

\textbf{The LFW face database} \cite{huang2007labeled}: Besides face images under controlled conditions, our model is also evaluated in the wild. In our experiment, we train on CelebA dataset \cite{liu2015deep} and test on LFW dataset \cite{huang2007labeled}. CelebA dataset consists of 202,559 celebrity face images. Each image is annotated with 5 landmarks. LFW dataset is a standard test set for verification in unconstrained conditions, which contains 13,233 images of 5,749 people. Following the verification protocol \cite{huang2007labeled}, 6,000 face pairs with 3,000 positive pairs and 3,000 negative pairs are provided to evaluate our model. 

\subsection{Baselines and Implementation Details}
\begin{figure*}[htb]
    \centering 
    \subfigure[AR dataset]{\includegraphics[width=0.26\textwidth]{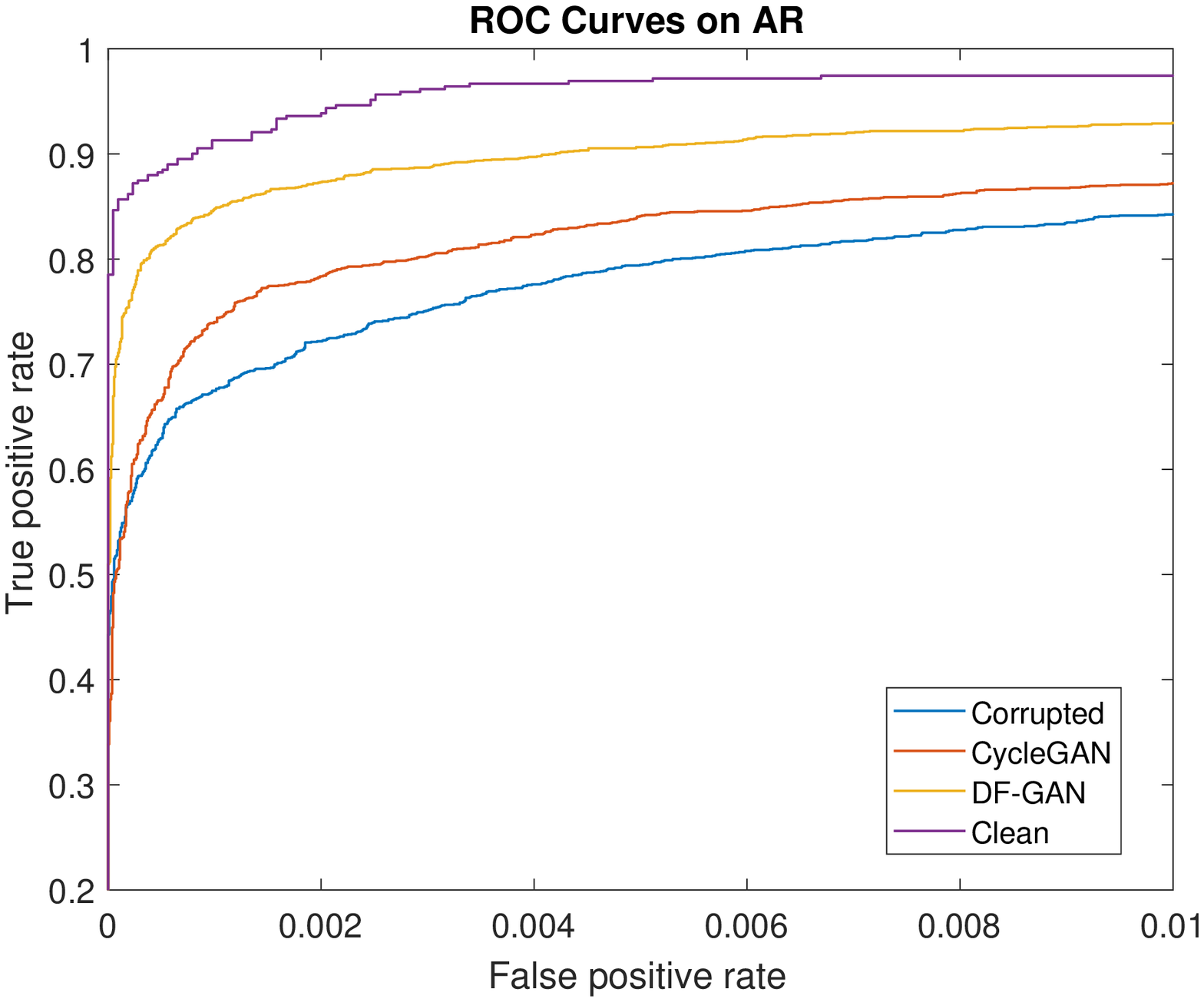}} \hspace{-1.5em}
    \subfigure[MultiPIE dataset]{\includegraphics[width=0.26\textwidth]{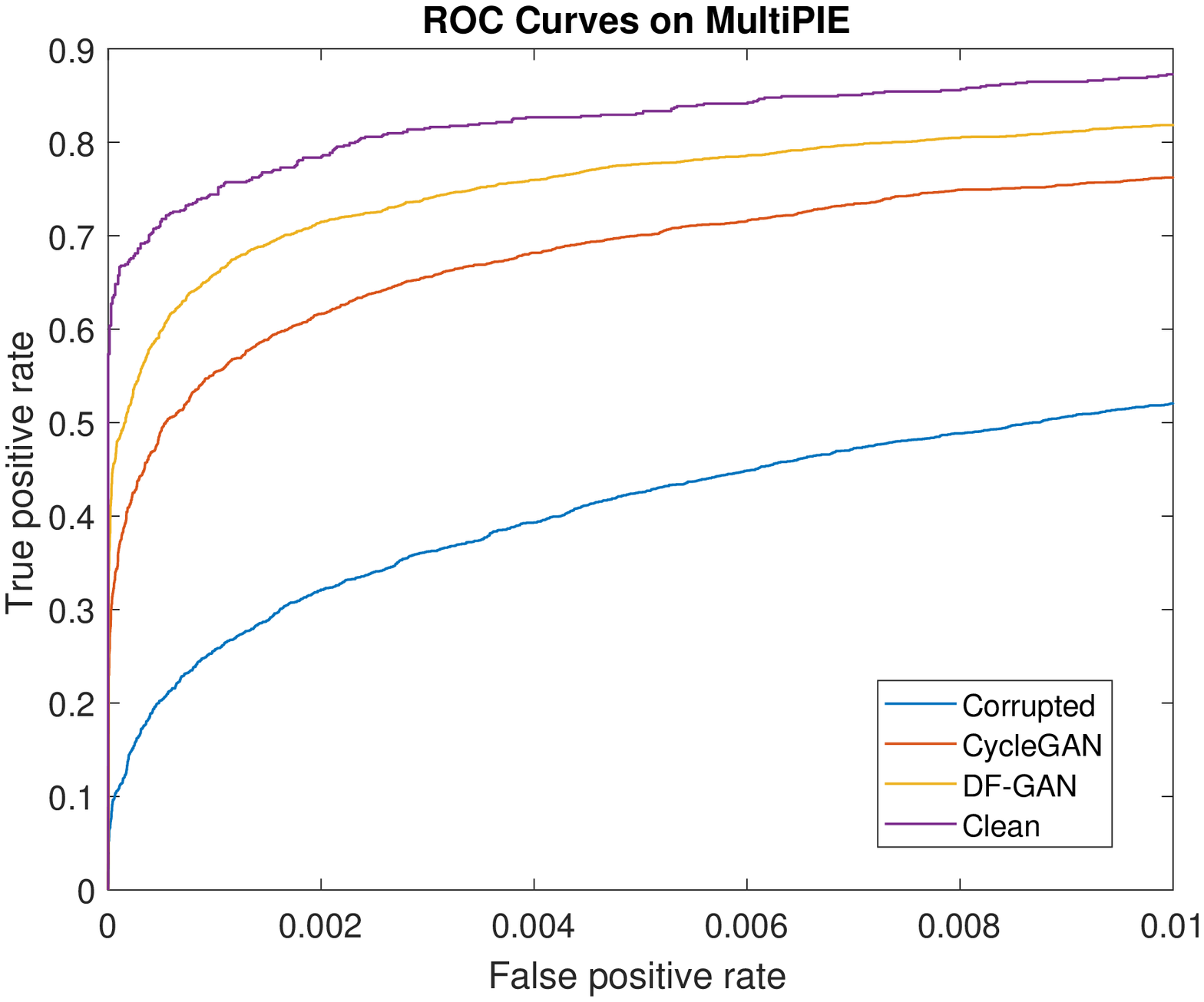}} \hspace{-1.5em}
    \subfigure[Color FERET dataset]{\includegraphics[width=0.26\textwidth]{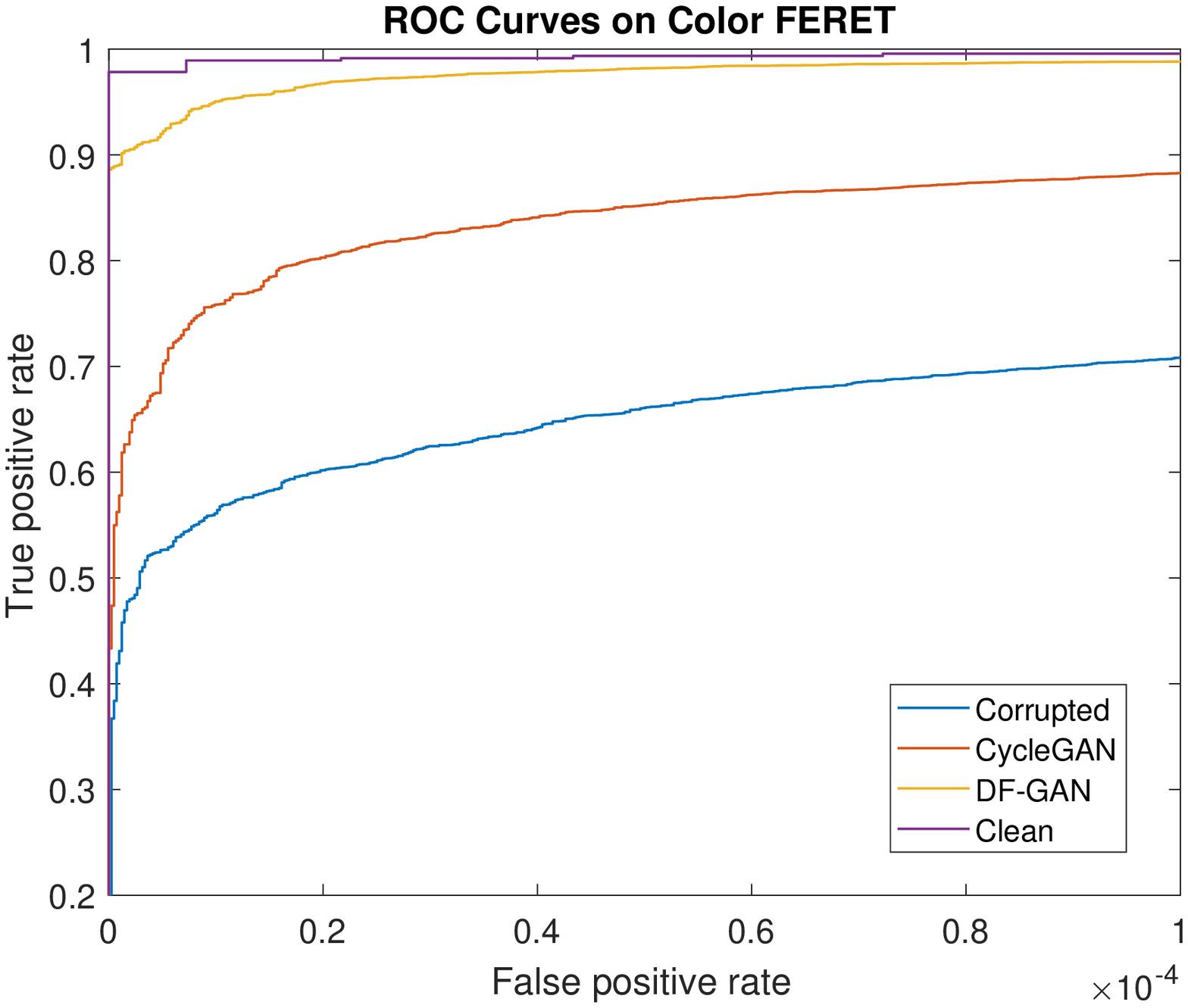}} \hspace{-1.5em}
    \subfigure[LFW dataset]{\includegraphics[width=0.26\textwidth]{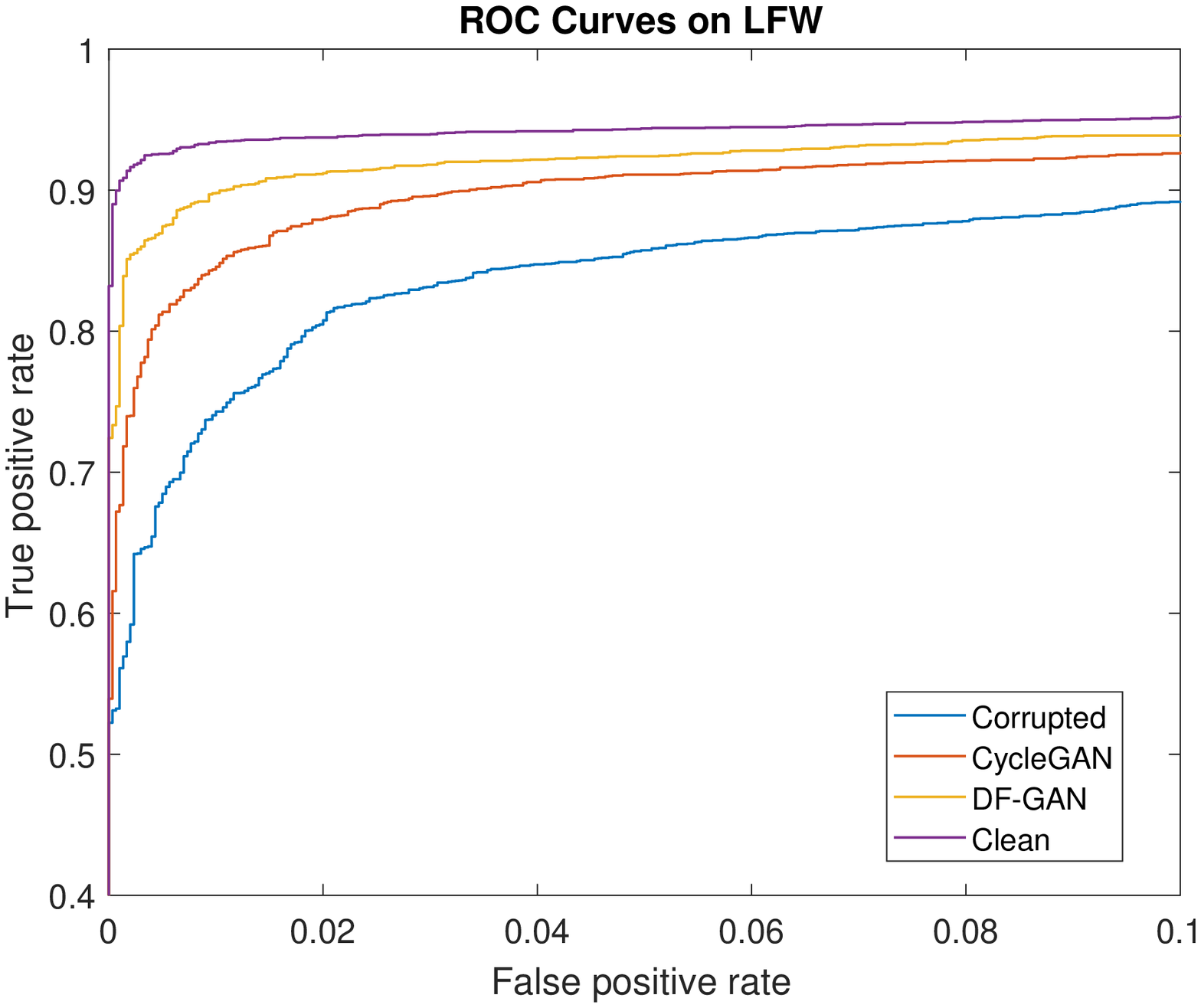}}
    \caption{ROC curves for AR, MultiPIE, Color FERET and LFW dataset.}
     \label{fig::roc}
\end{figure*}
Since large scale MeshFace images are difficult to obtain, we follow \cite{zhang2016multi} to generate the MeshFace images by adding random patterns to clean faces. Firstly, the completely random binary patterns are synthesized with differently random magnitudes and phases. Then, Gaussian filtering is adopted to smooth the binary patterns to get mesh images, which is corresponding to domain $Z$ in our method. Finally, the mesh is added to the clean face to generate MeshFace $x$ with the following formula:
\begin{equation}
x = \left\{ {\begin{array}{*{20}{c}}
{\beta M + (1 - \beta )y}&{M < 1}\\
y&{M = 1}
\end{array}} \right.
\end{equation}
Where $M$ is the mesh image in domain $Z$ and $y$ is the clean face. The parameter $\beta $ controls the transparency of occlusions. Similar to \cite{zhang2016multi}, each clean face is used to synthesize 30 totally random MeshFaces on the AR, MultiPIE and Color FERET databases. Due to memory limitations, we only synthesize one random MeshFace for each clean face on CelebA and LFW databases.

In the testing phase, the frontal view and neutral expression face of each individual is selected as a gallery set and the rest faces are used as a probe set on the AR, MultiPIE and Color FERET datasets. One image of each pair on LFW is randomly chosen and corrupted as a MeshFace. The face verification performance between the gallery set and recovered faces is compared through visual and quantitative analysis. We utilize TPR@FPR=$1\%$ (true positive rate when false positive rate is  $1\%$), TPR@FPR=$0.1\%$ and TPR@FPR=$0.01\%$ as evaluation criteria of face verification. The PSNR [dB] and SSIM \cite{wang2004image} are also reported to evaluate the quality of generated images.

Our DF-GAN is able to disentangle and fuse clean faces and occlusions to realize the completion with unpaired data. Although \cite{zhang2016multi,zhang2016demeshnet} present their methods to recover clean faces from MeshFaces, they need to know the position of occlusions for training. Moreover, the paried clean faces are required for the recently proposed face completion methods \cite{zhang2016multi,zhang2016demeshnet,GFC-CVPR-2017, yeh2017semantic}. Thus, our method does not compare with them. Since CycleGAN achieves the transformations between two domains with unpaired data, we take it as a baseline in our experiment, where CycleGAN acts on the domain $X$ and domain $Y$. Furthermore, face verifications using MeshFaces (Corrupted) and clean faces (Clean) directly are taken as a lower and a upper bound, respectively. 

All the face images are normalized using the positions of two eyes, and then resized to 148x148. Horizontal mirroring with 0.5 probability and random crops into 128x128 are used for data augmentation. In our experiments, all networks are trained using ADAM solver \cite{kingma2014adam} with batch size 16. An initial learning rate is set to 0.0001 for the first 50 epochs and linearly decays over the next 50 epochs. The trade-off parameter $\lambda $ is assigned to 10.

Our disentangling network $G$ and fusing network $F$ take the encoder-decoder architecture. The encoders and decoders for domain $X$, domain $Y$ and domain $Z$ consist of 5 residual blocks, 5 residual blocks and 1 residual block, respectively. We use fully convolution layers without any fully connected layers in our networks. Specifically, the size of the disentangled latent representation for MeshFace is 256x32x32, including 255x32x32 feature representation for the clean face and 1x32x32 feature representation for the occlusion. All experiments are conducted with PyTorch framework on a single GTX Titan X GPU.

\subsection{Evaluation of Verification Results}

The goal of DF-GAN is not only to generate clean faces, but also to preserve the identity information and improve the verification performance. In order to test to what extent the face identity can be preserved, our approach is evaluated on four datasets, i.e., AR, MultiPIE, Color FERET and LFW datasets. We firstly extract deep features with Light-CNN \cite{wu2015light} and then compare the TPR@FPR with a consine-distance metric. The recovered faces from MeshFaces by different methods are employed on face verification task according to the aforementioned testing protocol. Moreover, the verification performances using MeshFaces are taken as a fair comparison.

\begin{figure*}[htb]
   \vspace{-3pt}
    \centering
    \includegraphics[width=1\textwidth]{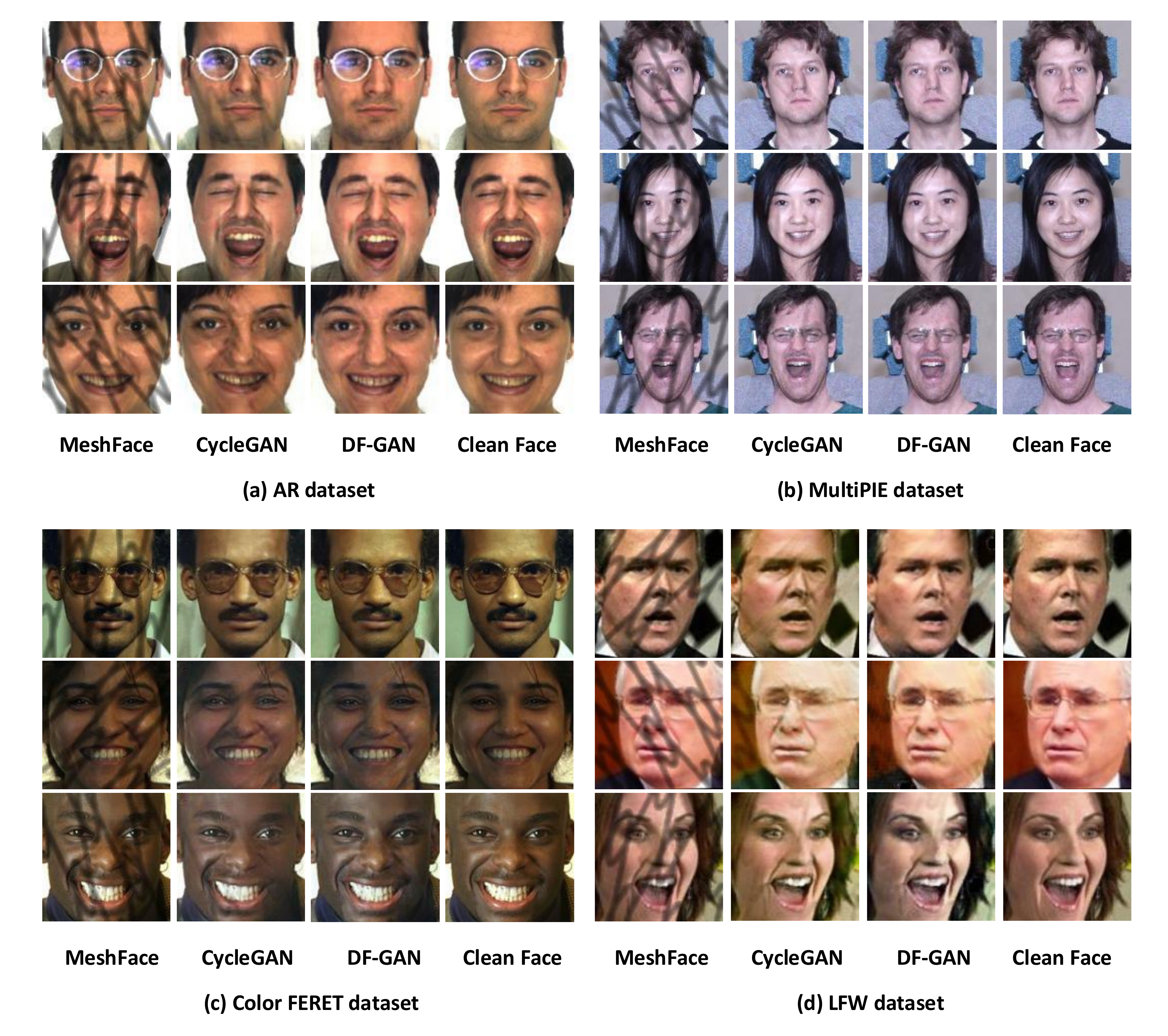}
    \caption{A visual comparison of the completion results on the test set of AR, MultiPIE, Color FERET and LFW dataset.}
     \label{fig::images}
     \vspace{-3pt}
\end{figure*}
As shown in Table \ref{tab::TPR_FPR}, directly using MeshFaces for verification, the preformance on all database declines sharply, which proves the fact that MeshFaces damage the face verification system greatly. Obviously, when MeshFaces are processed by face completion models, the performance of face verification is significantly improved. The underlying reason may be that face completion models push the distribution of recovered faces towards the real distribution of clean faces.

Experiments on AR, MultiPIE and Color FERET datasets are under highly controlled condition, so the results on these three datasets outperform that on LFW dataset. From Table \ref{tab::TPR_FPR}, it is obvious that using the recovered faces instead of MeshFaces for verification enormously enhances the performance. Especially on MultiPIE dataset, the performance of CycleGAN has improvement about 40\% at TPR@FPR=1\%, 100\% at TPR@FPR=0.1\% and 300\% at TPR@FPR=0.01\% comparing with the method using occluded faces directly. Our proposed DF-GAN achieves state-of-the-art results on all datasets, almost improving 60\% at TPR@FPR=1\%, doubling at TPR@FPR=0.1\% and even five times at TPR@FPR=0.01\% on MultiPIE dataset. Comparing DF-GAN with CycleGAN, DF-GAN surpasses about 6\% at TPR@FPR=1\%, 11\% at TPR@FPR=0.1\% and 20\% at TPR@FPR=0.01\% on AR dataset. This can be attributed to that DF-GAN explicitly models the distribution of structured occlusions that is helpful to face completion, while CycleGAN just utilizes the information of two domains. Moreover, it is worth noting that the performance on clean faces is an upper boundary of our model. Although our model is trained with unpaired data, the gap between DF-GAN and Clean at TPR@FPR=1\% is very small (about 4\% on AR dataset, 6\% on MultiPIE dataset and even reaches the upper limit on Color FERET dataset), which confirms the effectiveness of the proposed DF-GAN.

With the exception of experiments under controlled conditions, we also conduct an experiment on LFW dataset in the wild to test the generalization ability of the proposed DF-GAN. Specifically, the CelebA dataset is taken as training set and the LFW dataset is taken as testing set since images on both datasets are taken under unconstrained condition. Table \ref{tab::TPR_FPR} shows that even though the performance on LFW dataset is a little lower than that on AR and Color FERET dataset, DF-GAN achieves about 20\% improvement at TPR@FPR=1\%, 60\% improvement at both TPR@FPR=0.1\% and TPR@FPR=0.01\%. Besides, the gap between DF-GAN and the upper boundary is small, which demonstrates that our method achieves considerable progress in the face completion problem. Finally, we also plot the ROC curve for all datasets in Figure \ref{fig::roc} to validate the effectiveness of the proposed DF-GAN. 

\subsection{MeshFace Completion and Ablation study}

In this section, extensive quantitative and visual evaluations of the completion results are conducted on AR, MultiPIE, Color FERET and LFW datasets. For MeshFace completion, the visual results and quantitative evaluations of DF-GAN and the compared methods are reported. Then, we do ablation studies to investigate the capability of occlusion modeling of our proposed DF-GAN.

\subsubsection{MeshFace Completion}

MeshFace completion can be seen as an image transformation problem with identity maintained. Based on the CycleGAN \cite{CycleGAN2017} that is only able to handle the transformations between two domains, DF-GAN is capable of separating and reconstructing the clean face and the occlusion by adversarial training on three domains. According to Figure \ref{fig::images}, most of the occluded areas in MeshFaces are recovered after being processed by face completion model. In comparison, DF-GAN generates cleaner faces than CycleGAN on all databases. To evaluate the robustness of our approach, our databases contain different expressions (on the AR and MultiPIE datasets) and illumination conditions (on the Color FERET dataset). It is obvious that DF-GAN is invariant to expressions and illumination variations. In addition, experiment on LFW dataset is employed to evaluate the generalization capability of our model in the wild. Although the LFW dataset contains lots of low-resolution and large-pose variance images, DF-GAN can still recover the occluded areas completely.

Furthermore, the metrics PSNR and SSIM \cite{wang2004image} are used to quantitatively evaluate the recovered faces quality, where higher values of PSNR and SSIM indicate better results. As shown in Table \ref{tab::PSNR_SSIM}, the quantitative results are in consistent with our visual perception.

\subsubsection{Ablation Study}
To investigate the effectiveness of occlusion modeling of our proposed DF-GAN, we evaluate our method on two tasks: mesh separation and generation.

\begin{figure}[htb]
\vspace{-3pt}
    \centering
    \includegraphics[width=0.48\textwidth]{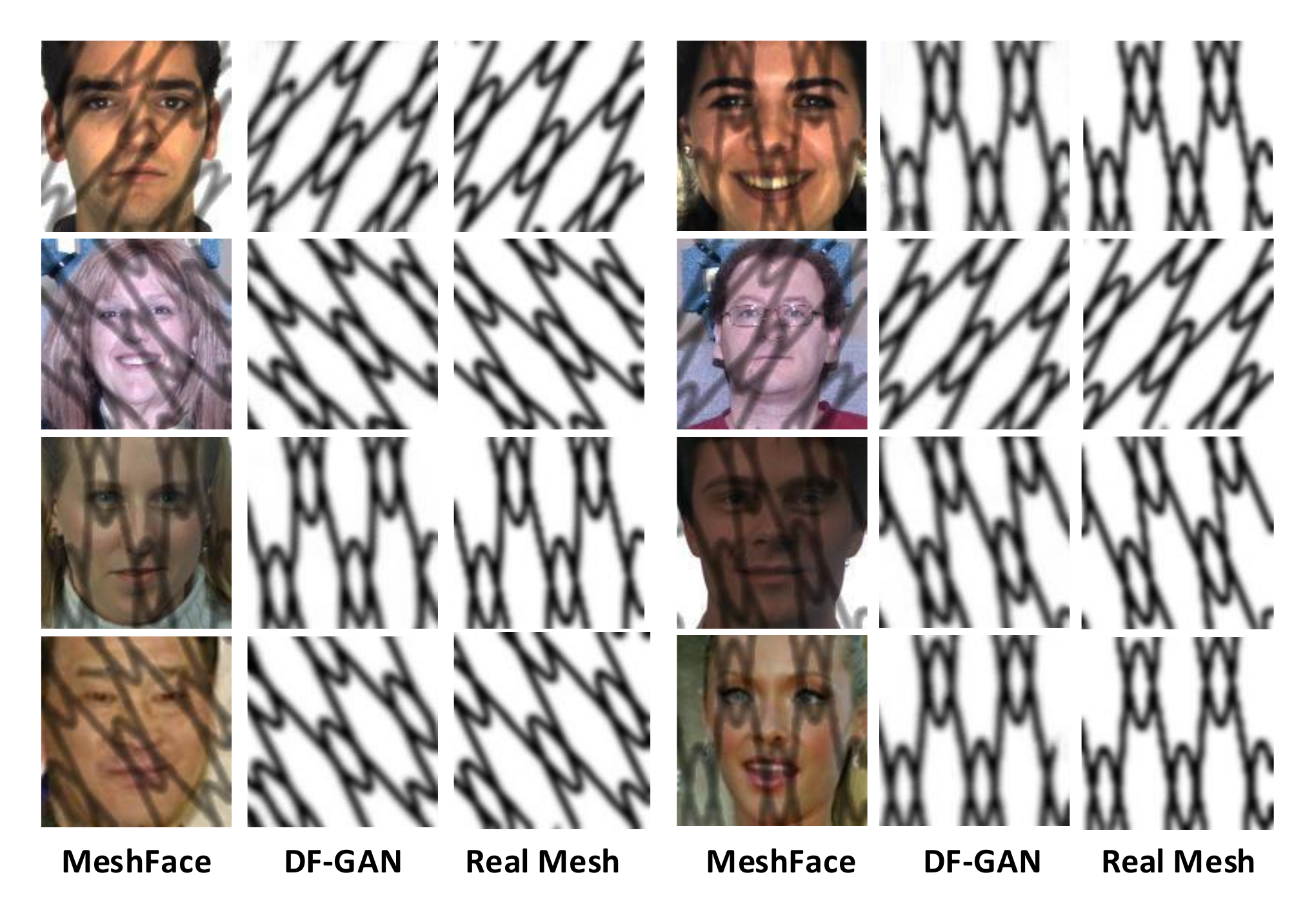}
    \caption{Visual inspection of mesh separation on four datasets.}
     \label{fig::add_mesh1}
     \vspace{-3pt}
\end{figure}

\noindent \textbf{Mesh separation in MeshFaces}: The disentangling network is able to separate mesh from a MeshFace. It is worth noting that the mesh can be an arbitrary random pattern that has never occurred in the training set. Figure \ref{fig::add_mesh1} shows some visual results of DF-GAN. It can be observed that DF-GAN is able to separate photorealistic random patterns. This experiment demonstrates that DF-GAN has learned the distribution of the random patterns instead of just remembering those random patterns in training set.

\begin{figure}[htb]
\vspace{-3pt}
    \centering
    \includegraphics[width=0.48\textwidth]{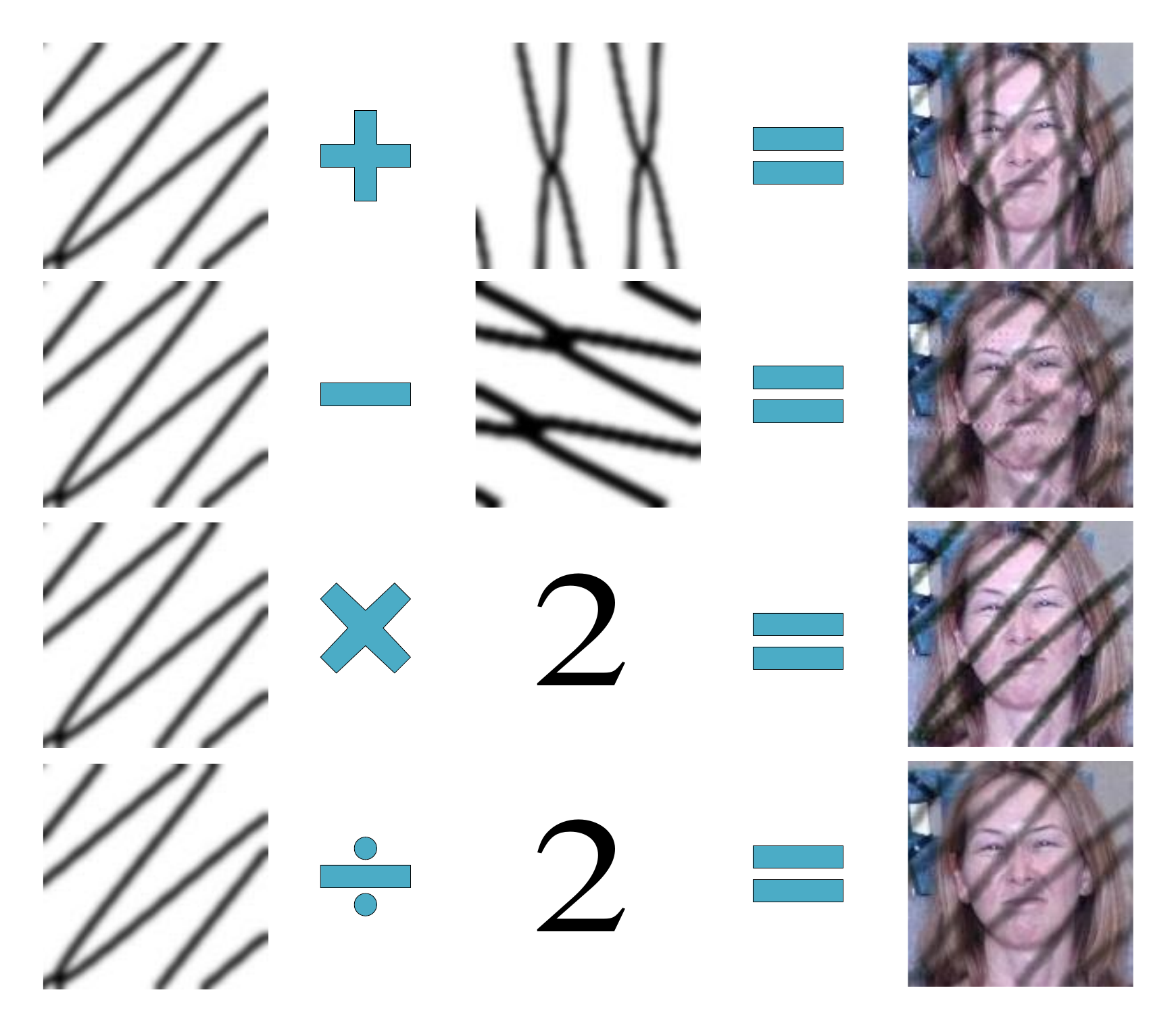}
    \caption{Visual inspection of MeshFace generation by vector arithmetic. It is worth noting that mesh images denote the corresponding feature representations.}
     \label{fig::add_mesh2}
     \vspace{-3pt}
\end{figure}

\noindent \textbf{Mesh generation by vector arithmetic of meshes}: \cite{mikolov2013distributed} has demonstrated that simple arithmetic operations reveal rich linear structure in representation space. We investigate whether similar structure emerges in mesh representation. Taking two meshes ${z_1},{z_2}$, we extract features ${F_{Enc - Z}}({z_1}),{F_{Enc - Z}}({z_2})$ from ${F_{Enc - Z}}$. The interpolation between ${F_{Enc - Z}}({z_1})$ and ${F_{Enc - Z}}({z_2})$ is capable of generating different representations, which are concatenated with a representation of clean face and fed into ${F_{Dec - X}}$ to synthesize the MeshFace. As shown in Figure \ref{fig::add_mesh2}, the superimposed image is sythesized by adding two representations of mesh images. Interestingly, when the feature vector of a mesh subtracts the other one, the cross regions of the two meshes are not occluded. Moreover, the numerical scaling of the feature representation controls the transparency of mesh. This indicates that structured occlusions have been effectively modeled by DF-GAN.

\section{Conclusion}

In this paper, we have developed a simultaneously disentangling and fusing Generative Adversarial Network for the face completion under structured occlusions. The transformations between the three domains are learned by the disentangling and fusing networks with an encoder-decoder structure, which allows simulating structure occlusions by modifying the latent representations. The proposed DF-GAN explicitly learns the distribution of occlusions, which is beneficial for face completion. Particularly, the paired face images are not required for training. Experimental results on four Meshface databases show that our model not only generates plausible visual completion results, but also improves the face verification performance. While the application example focuses on face in this paper, our DF-GAN is general and easily extended to style transfer such as zebras and horses, which is the focus of our future work.

{\small
\bibliographystyle{ieee}
\bibliography{egbib}
}

\end{document}